\documentclass[11pt]{article}

\usepackage[final]{acl}

\usepackage{times}
\usepackage{latexsym}

\usepackage{booktabs}
\usepackage{amsfonts}
\usepackage{multirow}
\usepackage{amsmath}
\usepackage{longtable}

\usepackage[T1]{fontenc}

\usepackage[utf8]{inputenc}

\usepackage{microtype}

\usepackage{inconsolata}

\usepackage{graphicx}

\title{Detecting Hallucinations in Large Language Models via Internal Attention Divergence Signals}

\author{Gijs van Dijk \\
  Utrecht University \\
  \texttt{g.vandijk1@students.uu.nl}}

\begin{document}
\maketitle
\begin{abstract}
We propose a lightweight and single-pass uncertainty quantification method for detecting hallucinations in Large Language Models. The method uses attention matrices to estimate uncertainty without requiring repeated sampling or external models. Specifically, we measure the Kullback–Leibler divergence between each attention head’s distribution and a uniform reference distribution, and use these features in a logistic regression probe. Across multiple datasets, task types, and model families, attention divergence is highly predictive of answer correctness and performs competitively with existing uncertainty estimation methods. We find that this signal is concentrated in middle layers and on factual tokens such as named entities and numbers, suggesting that attention dynamics provides an efficient and interpretable white-box signal of model uncertainty.
\end{abstract}

\section{Introduction}

LLMs achieve strong performance across tasks such as question answering, summarization, and reasoning \cite{zhao-2023}. Despite these advances, LLMs are known to generate incorrect or unsupported content, often referred to as hallucinations \citep{kalai-2025}, which makes them less reliable in situations where factuality is important. An issue with addressing this problem is that model outputs typically do not reflect the model's own uncertainty. Autoregressive language models are trained to generate fluent continuations, which can result in overconfident falsehoods. As a result, users cannot reliably distinguish between correct and incorrect outputs.

Hallucinations pose large risks in high-stakes areas such as healthcare \cite{kim-2025}, law \cite{magesh-2024}. Prior work has emphasized the need to detect and mitigate hallucinations, particularly in settings requiring factual reliability \citep{huang-2024}. Existing methods often conflate fluency with correctness or require substantial computational overhead \citep{liu-2025}.

We propose a simple, single-pass uncertainty measure derived from attention distributions. We compute the Kullback–Leibler (KL) divergence between each attention head’s distribution and a uniform reference distribution representing maximum uncertainty. Intuitively, reliable knowledge may correspond to concentrated attention on informative context tokens, whereas epistemic uncertainty may manifest as diffuse or misallocated attention. We aggregate these divergence signals across heads and layers and use a lightweight lasso-regularized probe to predict answer correctness. The probe serves only to aggregate signals across attention heads; the underlying uncertainty signal originates from the attention divergence itself.

Across multiple datasets, task types, and model families, attention divergence is highly predictive of answer correctness and performs competitively with existing uncertainty estimation methods. We find that the signal is concentrated in middle layers and peaks at factual tokens such as named entities and numbers, suggesting that internal attention dynamics provide an efficient white-box signal of model uncertainty.

\section{Background}
Recent work suggests that hallucinations in Large Language Models (LLMs) are not merely decoding errors, but arise from properties of the transformer architecture \citep{vaswani-2017} and its learned internal representations \citep{huang-2024, orgad-2024}

Uncertainty can be separated into aleatoric and epistemic uncertainty \citep{KIUREGHIAN2009105, hullermeier-2021}. In the context of natural language modelling, aleatoric uncertainty is often associated with ambiguous prompts, underspecified questions, or multiple equally valid continuations \citep{hou-2023, ling-2024}. This type of uncertainty is inherent to the input and cannot be reduced even with more training data.

Epistemic uncertainty, in contrast, captures uncertainty due to limited knowledge of the model, finite training data, model misspecification, or gaps in learned representations. Unlike aleatoric uncertainty, epistemic uncertainty is, in principle, reducible. This distinction is particularly important for hallucination detection in LLMs. Hallucinations characterized by untruthful outputs are not primarily driven by aleatoric uncertainty, as they typically do not arise from ambiguous prompts. Instead, they reflect epistemic failures where the model produces fluent output despite lacking reliable knowledge \citep{huang-2024}. In such cases, the model appears confident even when its internal knowledge is insufficient.

Uncertainty quantification provides a framework for addressing hallucinations in Large Language Models. Most existing approaches operate in the output space, estimating uncertainty based on the generated text or its associated probabilities. Common methods rely on logit- and likelihood-based signals such as log-likelihood, perplexity, maximum token probability, or predictive entropy \citep{liu-2025}. These approaches assume that low-probability generations correspond to higher uncertainty.

However, this assumption does not hold. Hallucinated statements can receive high likelihood under the model distribution, as autoregressive language models are trained to generate fluent continuations rather than calibrated confidence estimates. As a result, output-based uncertainty signals can fail to distinguish confident errors from reliable knowledge.

This motivates a shift toward internal signals. If hallucinations may arise from failures in internal computation, then signals extracted from hidden states, attention patterns, or other latent representations may provide more reliable indicators of epistemic uncertainty. In this work, we focus on attention as a structured internal probability distribution that may encode such signals.

\section{Related Work}

A growing body of research has investigated hallucination detection and uncertainty quantification using internal model signals rather than only output probabilities.

\subsection{Hidden State Probing}
\citet{orgad-2024} show that internal representations encode signals of truthfulness concentrated on answer tokens. Similarly, \citet{binkowski-2025} and \citet{chen-2024} train probes over hidden states to detect hallucinations. 
However, these approaches often struggle to generalize across tasks and datasets. Probes trained on hidden states tend to capture task-specific correlations rather than a global confidence signal.

\subsection{Attention-based Methods}

Several recent studies have explored methods that use attention to infer uncertainty signals. \citet{li-2025} introduce Uncertainty Quantification with Attention Chain (UQAC), a white-box method that uses attention weights to identify which reasoning tokens are most influential for producing the answer. Similarly, TOHA (Topology-based Hallucination detector) \citep{bazarova-2025} analyses topological properties of attention matrices to estimate uncertainty.

In the supervised setting, methods such as Lookback Lens \citep{chuang-2024} and Attention-Pooling Probes \citep{ch-wang-etal-2024-androids} train lightweight classifiers over attention-derived features. Lookback Lens uses per-head ratios of attention to context versus generated tokens, while Attention-Pooling Probes pool attention weights across heads and layers. Our method instead uses a KL-divergence-based attention measure with a lightweight probe, relying on low-dimensional, interpretable features.

Other work shows that hallucinations might arise in specific attention heads. \citet{vazhentsev-2025} propose an attention-based uncertainty estimation approach by identifying a subset of uncertainty-aware attention heads whose behaviour changes at hallucinated tokens. \citet{stolfo-2024} similarly report certainty related signals localized in particular neurons.

\subsection{Sampling- and Output-Based Methods}

A large class of approaches estimates uncertainty directly from model outputs. Sampling-based methods measure disagreement across multiple generations, such as semantic entropy \citep{farquhar-2024} or SelfCheckGPT \citep{manakul-2023}. Other methods rely on output probabilities, perplexity, or entropy \citep{ren-2022, liu-2025}. Some approaches use external verifiers or ensembles \citep{kuhn-2023}.

While often effective, these methods typically require repeated sampling or additional models. In contrast, our method operates in a single forward pass and extracts uncertainty signals directly from attention distributions.

\section{Methodology}

\subsection{Kullback–Leibler Divergence}

To create a quantitative signal from attention, we measure how much the attention distribution deviates from a uniform baseline ($\mathcal{U}$) using Kullback–Leibler (KL) divergence.



It is often useful to quantify how well one probability distribution approximates another. We therefore use the Kullback--Leibler (KL) divergence, which measures the discrepancy between two distributions. Given a reference distribution $P$ and an alternative distribution $Q$, the KL divergence is defined as the expected log-ratio between $P$ and $Q$ under $P$.

In our setting, attention weights define a discrete probability distribution over token positions $x \in \{1,\dots,T\}$. The KL divergence therefore takes the form

\begin{equation}
    D_{\mathrm{KL}}(P \,\|\, Q)
    = \mathbb{E}_{x \sim P}\!\left[\log \frac{P(x)}{Q(x)}\right].
\end{equation}

Intuitively, this quantifies how much extra information (nats\footnote{Nats denote information measured using natural logarithms (base e).}) is required, on average, when outcomes drawn from distribution $P$ are interpreted as if they were drawn from $Q$. 

When the reference distribution $Q$ is uniform (we will denote this as $\mathcal{U}$ for clarity), the KL divergence simplifies to

\begin{equation}
    \sum_{x} P(x)\log\frac{P(x)}{1/T}
    = \log T - H(P)
\end{equation}

where $T$ denotes the number of tokens in the attention context window, so that $\mathcal{U}(x)={1}/{T}$ assigns equal probability to each context token position.

\begin{figure}[t]
  \centering
  \includegraphics[width=0.9\linewidth]{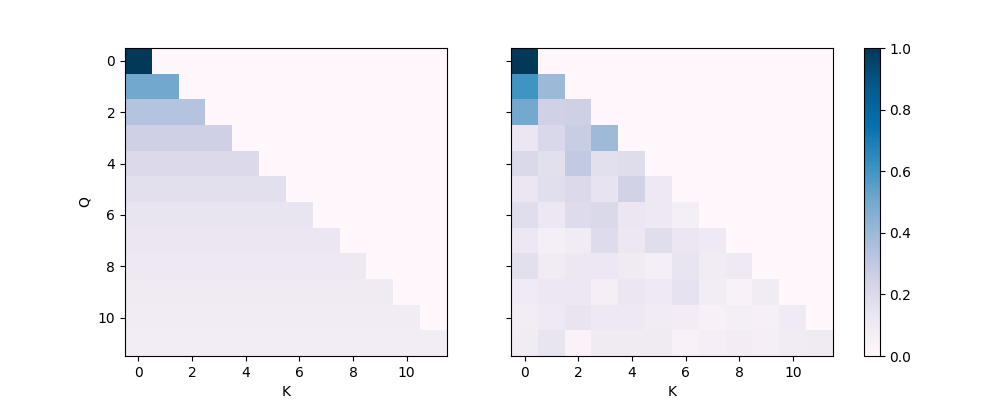}
  \caption{Intuition of attention patterns with low KL divergence to uniform (left)
and higher KL divergence to uniform (right). Higher divergence corresponds
to more concentrated attention.
}
  \label{fig:attention-diff-v-unif}
\end{figure}


A higher concentration of attention reflects stronger model confidence, not necessarily correctness. In our hypothesis, hallucinations arise when the model exhibits wrongly calibrated confidence, characterized by highly concentrated attention on misleading tokens. In contrast, near uniform attention distributions are more likely to occur at positions where the model lacks reliable knowledge and remains uncertain, since the model does not know to which token to attend.

\subsection{Attention Divergence}

In autoregressive transformer models, such as GPT, \citep{vaswani-2017}
the attention weights produced by a single attention head at a given generation step form a discrete probability distribution over the available context tokens. 
Let $A_t^{(l,h)}$ denote the attention distribution of head $h$ in layer $l$ at generation step $t$, defined over the $t$ previously generated tokens. We compare this distribution to a uniform reference distribution $\mathcal{U}$, which assigns equal probability to all context positions and represents a state of maximal uncertainty.

We quantify the separation between these two distributions using KL divergence. This measures how much an attention head focuses on a subset of previous tokens. For each attention head, the KL divergence is computed at every generation step and averaged across the generation answer tokens, yielding a single scalar feature per head. These per-head divergence values form a feature vector that summarizes the attention during generation.

Formally, for each example we construct a feature vector $x_i \in \mathbb{R}^{L \times H}$ where each entry corresponds to the mean KL divergence of a single attention head pooled over the generated answer tokens.

\subsection{Probing}

To predict answer correctness and uncertainty from the attention divergence features, we train a logistic regression probe with lasso (L1) regularization. The model estimates $P(y_i =1 \mid x_i) =\sigma(w^\top x_i+b)$, where $\sigma$ is the logistic sigmoid function, and $w \in \mathbb{R}^{L \times H}$ is the weight vector.

We train a logistic regression probe with L1 (lasso) regularization by minimizing
\begin{equation}
\begin{aligned}
\mathcal{L}(w,b)
&= -\sum_{i=1}^N \bigl[
y_i \log p_i + (1-y_i)\log(1-p_i)
\bigr] \\
&\quad + \lambda \lVert w \rVert_1
\end{aligned}
\end{equation}

where $p_i = \sigma(w^\top x_i + b)$, and $\lambda$ controls the strength of the sparsity penalty. We use lasso (L1) regularization to select a sparse subset of attention heads that the probe thinks are the most predictive of correctness. This provides insight into where these heads are located.

We use stratified k-fold cross-validation to assess stability. We report AUROC as the main metric, which is insensitive to class imbalance \citep{li2024arearoccurve} and captures the quality of probabilistic ranking. We treat correct answers as the positive class $(y=1)$. For each example $i$, the probe outputs a score $p_i=P(y_i=1\mid x_i)$ where $x_i$ denotes our attention divergence feature on the $i$-th generated answer, and $y_i\in \{0,1\}$ classifies whether that answer is correct. We compute AUROC over all examples (both correct and incorrect) by varying a threshold on $p_i$ and measuring the resulting true-positive and false-positive rates. Under this definition AUROC corresponds to the probability that a randomly chosen correct answer is assigned a higher score than a randomly chosen incorrect answer.

We additionally report the Expected Calibration Error (ECE) to see how well our probe is calibrated \citep{pavlovic-2025, wang-2023}. In addition, we measure accuracy. Although accuracy is not a reliable metric for class imbalance \citep{Kubat97addressingthe}, we still report it, as it provides additional context along AUC and ECE. To assess stability, experiments are repeated across multiple random dataset shuffles. 

\begin{table}[t]
\centering
\small
\setlength{\tabcolsep}{4pt}
\renewcommand{\arraystretch}{1.1}

\begin{tabular}{llc}
\toprule
Dataset & Model & AUROC \\
\midrule

TruthfulQA
& Llama-3.2-3B & $0.906 \pm 0.024$ \\
& Qwen3-4B     & $\mathbf{0.906 \pm 0.025}$ \\
& Mistral-7B   & $0.891 \pm 0.024$ \\
\addlinespace

TriviaQA
& Llama-3.2-3B & $0.835 \pm 0.004$ \\
& Qwen3-4B     & $\mathbf{0.846 \pm 0.001}$ \\
& Mistral-7B   & $0.835 \pm 0.004$ \\
\addlinespace

HotpotQA
& Llama-3.2-3B & $\mathbf{0.8035 \pm 0.0219}$ \\
& Qwen3-4B     & $0.7963 \pm 0.0217$ \\
& Mistral-7B   & $0.7782 \pm 0.0061$ \\
\addlinespace

GSM8K
& Llama-3.2-3B & $0.7672 \pm 0.0305$ \\
& Qwen3-4B     & $\mathbf{0.9450 \pm 0.0151}$ \\
& Mistral-7B   & $0.7889 \pm 0.0270$ \\

\bottomrule
\end{tabular}

\caption{
Validation results after training a lasso-regularized probe to predict correctness from attention divergence mean pooled across the generation.
Within each dataset, the highest AUROC is shown in bold.
Results are reported as mean $\pm$ standard deviation over three random seeds and five stratified cross-validation folds.
}
\label{tab:main_results_short}
\end{table}

\section{Experiments}
\subsection{Configuration}
For our main evaluation we use three different instruction tuned models: Llama-3.2-3B-Instruct \citep{grattafiori-2024}, Qwen3-4B-Instruct \citep{yang-2025-qwen}, and Mistral-7B-Instruct-v0.2 \citep{jiang-2023}. Models with instruction tuning create more structured outputs without requiring much prompt engineering \citep{zhang-2023}.

We evaluate our method on 4 different datasets spanning multiple task categories. TriviaQA \citep{joshi-2017} (open-domain) and TruthfulQA \citep{lin-2021} (multiple-choice) for factual question answering, HotpotQA \citep{yang-2018-hotpot} for multi-hop reasoning question answering, and GSM8K \citep{cobbe2021gsm8k} for mathematical reasoning. For each dataset, we sample data points from 3 different seeds, and use stratified k-fold cross-validation within sample. Answer correctness is computed using task-specific evaluation rules. For TriviaQA, TruthfulQA, and HotpotQA, answers are marked correct based on string matching against gold references. For GSM8K, correctness is computed by extracting and comparing the final numeric answer from the raw generated output. Full experimental details, including exact sample counts per dataset and per seed, are provided in the appendix.

\subsection{Results}

Table \ref{tab:main_results_short} summarizes the performance of the attention divergence signal across different datasets and models. The full results can be found in the appendix table ~\ref{tab:main_results} which includes accuracy and Expected Calibration Error (ECE).

Attention divergence is predictive of answer correctness across a range of tasks and model families. 
On factual questioning (QA) benchmarks (TruthfulQA and TriviaQA), performance is particularly strong. Across all three models, AUROC values exceed 0.89 on TruthfulQA and 0.83 on TriviaQA, with relatively small variance across seeds and folds. This suggests that our measure is stable for detecting incorrect or hallucinated answers. Despite accuracy not being the most reliable metric in situations with high class imbalance \citep{Kubat97addressingthe}, it has similar results, with values generally above 0.80 for TruthfulQA and close to or above 0.77 on TriviaQA. The ECE remains moderate on TruthfulQA, which comes partly due to the smaller sample size of just 216 samples across 3 seeds. The ECE is notably low on TriviaQA, indicating that the probe is reasonably well calibrated. 

For multi-hop reasoning on HotpotQA, performance is slightly lower. AUROC values range from 0.78 to 0.80. Accuracy remains in the 0.70 range, and calibration is comparable to TriviaQA, with most ECE values below 0.10. 

On mathematical reasoning (GSM8K), performance varies more strongly. Qwen3-4B-Instruct achieves high AUROC and accuracy, whereas Llama and Mistral show more moderate results. Despite this variability, AUROC values remain well above chance for all models. In a binary setting, an AUROC of 0.5 corresponds to random ranking, i.e., the model assigns higher scores to correct than incorrect answers only half the time on average. Even in generation tasks with many steps, the method still provides an uncertainty signal. Calibration on GSM8K is also good, with low ECE values (0.04-0.06) across all models. 

\begin{table}[t]
\centering
\small
\setlength{\tabcolsep}{6pt}
\renewcommand{\arraystretch}{1.1}

\begin{tabular}{lcc}
\toprule
Method & Single Gen. & Mistral-7B \\
\midrule
\textbf{Ours} & \textbf{Y} & $\mathbf{0.78 \pm 0.02}$ \\
\midrule
TOHA [\citenum{bazarova-2025}] & \textbf{Y} & $0.71 \pm 0.08$ \\
SelfCheckGPT [\citenum{manakul-2023}] & N & $0.70 \pm 0.06$ \\
Semantic entropy [\citenum{farquhar-2024}] & N & $0.70 \pm 0.05$ \\
EigenScore [\citenum{chen-2024}] & N & $0.68 \pm 0.04$ \\
HaloScope [\citenum{du-2024}] & \textbf{Y} & $0.60 \pm 0.06$ \\
LLM-Check [\citenum{NEURIPS2024_3c1e1fdf}] & \textbf{Y} & $0.48 \pm 0.03$ \\
Perplexity [\citenum{ren-2022}] & \textbf{Y} & $0.55 \pm 0.06$ \\
Max entropy [\citenum{fadeeva-2024}] & \textbf{Y} & $0.62 \pm 0.04$ \\
ReDEEP [\citenum{sun-2024}] & \textbf{Y} & $0.49 \pm 0.04$ \\
\bottomrule
\end{tabular}

\caption{
AUROC on the HotpotQA dataset using Mistral-7B.
Baseline results are taken from \citet{bazarova-2025}. Our method is evaluated under the same model and dataset.
}
\label{tab:hotpot_comparison}
\end{table}

\begin{table}[t]
\centering
\small
\setlength{\tabcolsep}{6pt}
\renewcommand{\arraystretch}{1.1}

\begin{tabular*}{\linewidth}{@{\extracolsep{\fill}}lccc}
\toprule
Method & TriviaQA & TruthfulQA \\
\midrule
\textbf{Ours} & ${0.835}$ & $\mathbf{0.906}$ \\
\midrule
LapEigvals [\citenum{binkowski-2025}] & $\underline{0.836}$ & $\underline{0.829}$ \\
Hidden States (probe) [\citenum{binkowski-2025}] & $\mathbf{0.850}$ & $0.823$ \\
\bottomrule
\end{tabular*}

\caption{
AUROC on TriviaQA and TruthfulQA using Llama-3.2-3B. The highest AUROC value is boldfaced, the runner-up is underlined.
}
\label{tab:Llama32_comparison}
\end{table}

\subsubsection{Comparison to Prior Work}

We compared our method to existing uncertainty quantification and hallucination detection methods in settings where direct comparison is possible. Table \ref{tab:hotpot_comparison} reports results on the HotpotQA \citep{yang-2018-hotpot} dataset using the Mistral-7B model \citep{jiang-2023}. In this setting, our method achieves an AUROC of $0.78 \pm 0.02$, outperforming both sampling-/ensemble-based methods and other single generation baselines reported by \citep{bazarova-2025}. Our method also improves over TOHA (\underline{To}pology-based \underline{Ha}llucination detector) \citep{bazarova-2025}, which is also attention-based and relies on topological features of attention matrices. Compared to output-based uncertainty measures, such as semantic entropy \citep{farquhar-2024}, SelfCheckGPT \citep{manakul-2023}, and other baselines \citep{ren-2022, sun-2024, fadeeva-2024, NEURIPS2024_3c1e1fdf, du-2024, chen-2024}, our method provides a stronger AUROC signal, while requiring only a single forward pass.

Table \ref{tab:Llama32_comparison} compares our results to \citet{binkowski-2025} on TriviaQA and TruthfulQA. Our method achieves competitive performance on TriviaQA and improves AUROC on TruthfulQA. Across both comparisons, our method consistently performs well, all while requiring less computation. These results support the claim that local attention dynamics encode meaningful uncertainty information and can be used for hallucination detection.

\subsection{Sanity Checks}
\label{sec:sanity_checks}

To verify that our proposed signal is not caused by possible other factors, we evaluate a set of sanity checks on TriviaQA using the Llama-3.2-3B-Instruct model. These baselines include simple properties of the generated output and prompt that could possibly correlate to answer correctness. Specifically, we measure: generation length, prompt length, raw output length, final token punctuation, and the amount of digits in the output. We calculate AUROC for all these baselines.

Each baseline is evaluated independently by computing its AUROC with respect to answer correctness. All sanity checks achieve AUROC values close to chance or exhibit weak reverse correlations with correctness.

Additionally, we compute AUROC after permuting the correctness labels. Because permutation destroys any true relationship between the features and the labels, we show that the measured signals are not caused by any leakage (such as the labels being known beforehand). These sanity check results show that our main results are not caused by any of these baselines.

The table for sanity checks can be found in the appendix.

\subsection{Ablation Experiments}
To better understand our measure we perform a series of ablation experiments on the TriviaQA dataset. The results are shown in Table ~\ref{tab:ablation_summary}.

First, we ablate the most influential attention heads that were identified by the L1-regularization probe by removing the top-$k$ heads ranked by absolute coefficient magnitude. Removing up to $k=50$ heads does not cause a drop in AUROC relative to the baseline on TriviaQA on Llama-3.2-
3B, it even slightly improves AUROC for some $k$ values. This indicates that the uncertainty signal is not per se localized to a small subset of heads, but is instead encoded across multiple correlated heads.

Next, we ablate entire groups of layers by removing early, middle, or late layers from the feature set. Removing early and middle layers leads to a reduction in AUROC, with the largest drop observed when removing middle layers. This suggests that the signal primarily depends on early-to-middle layers.

We also replaced mean pooling with max pooling. This resulted in a significant performance drop, indicating that the signal is not driven by a small subset of single attention spikes, but by consistent attention diffusion. 

\begin{table}[t]
\centering
\small
\setlength{\tabcolsep}{6pt}
\renewcommand{\arraystretch}{1.1}
\begin{tabular}{lc}
\toprule
Ablation & AUROC \\
\midrule
Remove top-5 heads & 0.857 $\downarrow$ \\
Remove top-10 heads & 0.853 $\downarrow$ \\
Remove top-20 heads & 0.862 $\uparrow$ \\
Remove top-50 heads & 0.872 $\uparrow$ \\
Remove early layers & 0.849 $\downarrow$ \\
Remove middle layers & 0.844 $\downarrow$ \\
Remove late layers & 0.862 $\uparrow$ \\
Max pooling & 0.795 $\downarrow$ \\
\bottomrule
\end{tabular}
\caption{Ablation results on TriviaQA using Llama-3.2-3B. Without ablations the result was an AUROC of 0.858. Increases in AUROC are annotated with $\uparrow$, decreases by $\downarrow$}
\label{tab:ablation_summary}
\end{table}

Additionally, we perform an ablation experiment where our feature is computed over three different subsets of the sequence. Specifically, we compute the signal over just the prompt tokens before any tokens are generated, the answer-only, where it is pooled over just the generated token, and the full prompt and answer combination. For solely the prompt tokens we achieve an AUROC of 0.7674, for the answer 0.8707, and for the full generation 0.8215. 

Since the prompt only condition is far above chance, a proportion of the uncertainty signal is already present before the model begins to generate tokens. This supports the idea that uncertainty can also be aleatoric, i.e., related to ambiguous prompts or lacking context.
This suggests that hallucinations are not solely the result of failures during output generation, but partially due to the prompt itself. 

The answer only condition yielded the strongest result, with an AUROC of 0.87. 

\subsection{Layer \& Head Analysis}

To investigate whether our proposed attention divergence measure is localized to a specific subset of layers and heads, we analyse the difference in mean attention divergence between correct and incorrect generations. We do this by plotting a heatmap with heads on the x-axis and layers on the y-axis. 

\begin{figure}[t]
  \centering
  \includegraphics[width=0.9\linewidth]{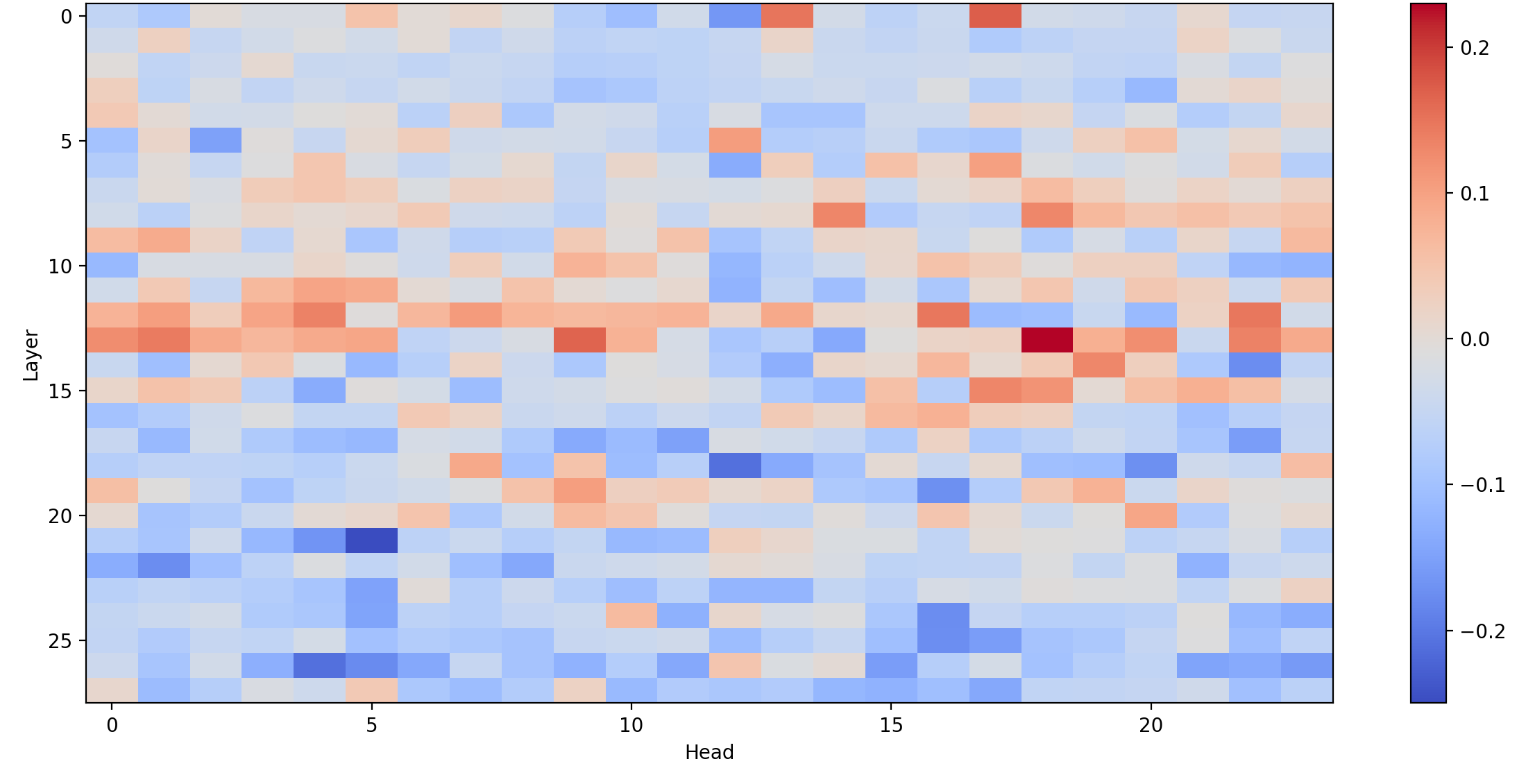}
  \caption{Heatmap of the difference in mean attention divergence between correct and incorrect generations, computed per layer and attention head. Positive values (red) indicate higher divergence for correct answers.}
  \label{fig:delta_kl_correct_miunus_incorrect}
\end{figure}

Figure ~\ref{fig:delta_kl_correct_miunus_incorrect} visualizes

\begin{equation}
\begin{aligned}
\Delta D_{\mathrm{KL}}(P \,\|\, \mathcal{U})
&= \mathbb{E}\!\left[
    D_{\mathrm{KL}} \mid \text{Correct}
\right] \\
&\quad - \mathbb{E}\!\left[
    D_{\mathrm{KL}} \mid \text{Incorrect}
\right]
\end{aligned}
\end{equation}

The strongest differences are concentrated in the middle layers of the model. This pattern suggests that our attention divergence measure is not uniformly distributed. Instead, it peaks at the middle layers and is distributed across multiple heads, rather than being dominated by a small subset of attention heads. This explains why removing individual heads has limited effect on overall performance.

We found that attention divergence differs between correct and incorrect generations. To better understand how extreme this difference is, we analyse the distributions using empirical cumulative distribution functions (ECDFs). This allows us to examine how the two groups differ across the entire distribution, with the tails in particular. 

\begin{figure}[t]
  \centering
  \includegraphics[width=0.9\linewidth]{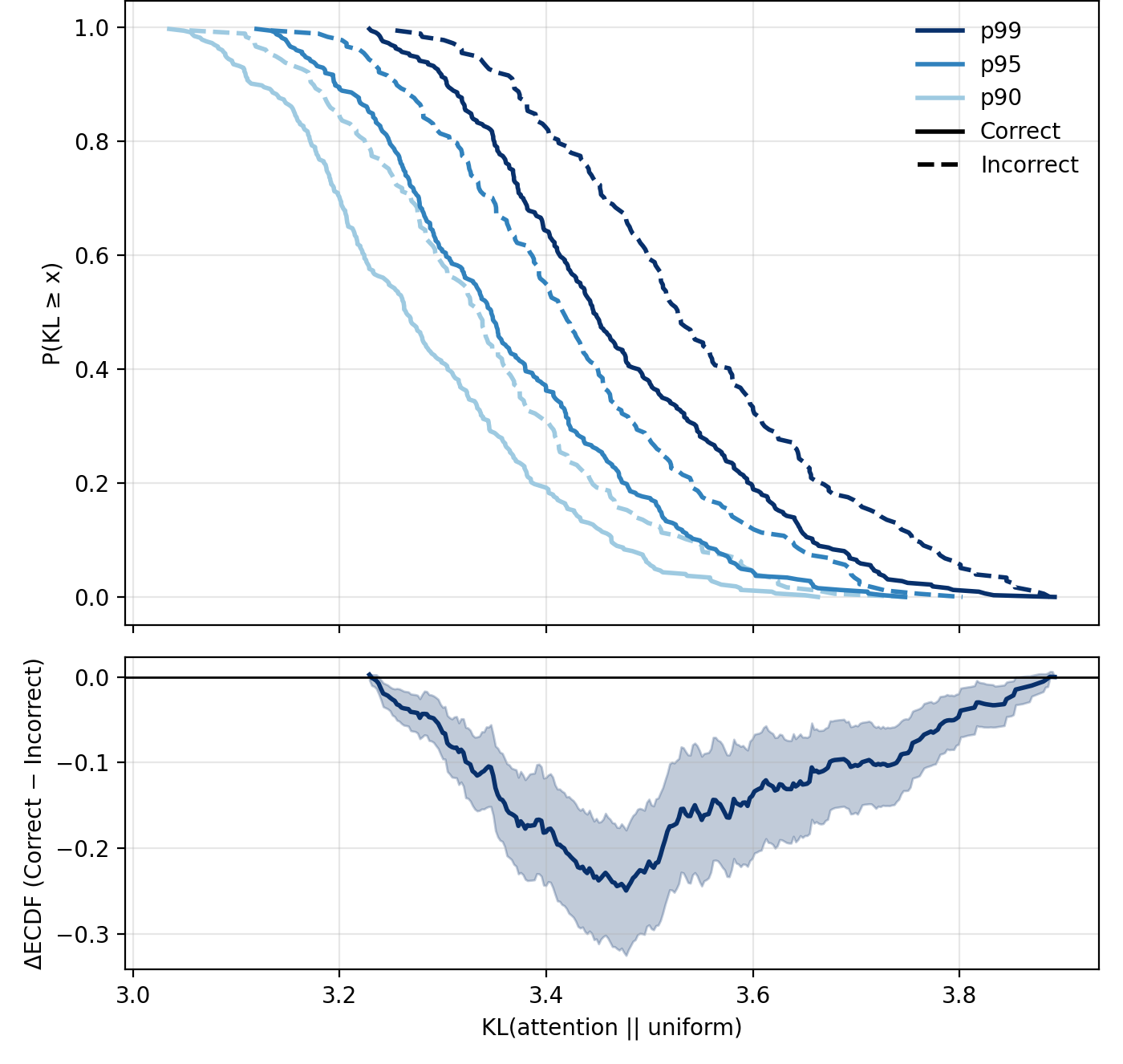}
 \caption{
    Empirical cumulative distribution functions (ECDFs) of attention divergence values for correct and incorrect generations.
    Top figure are survival ECDFs $P(\mathrm{KL} \geq x)$ computed at the $p \in \{90,95,99\}$ percentiles of the token-level KL divergence between attention distributions and a uniform baseline. Solid lines correspond to correct generations, dashed lines are incorrect generations.
    Bottom figure is the difference between the survival ECDFs for correct and incorrect generations at $p=99$, with shaded areas indicating confidence intervals (CI).
}
  \label{fig:ecdf}
\end{figure}

For each generated answer, we compute the $p\text{-th}$ percentile of the KL divergence between the attention distribution and a uniform baseline. The values for $p$ that we use are $p \in \{90,95,99\}$. The top section of Figure ~\ref{fig:ecdf} shows the ECDFs for $P(KL \geq x$), for correct and incorrect answers at each percentile level $p$. Solid lines correspond to correct generations, whereas dashed lines are incorrect generations. 

Across all percentile levels $p$, incorrect generations exhibit a rightward shift relative to correct generations, indicating more of a right tail in attention divergence. This separation increases along $p$, showing that the difference between correct and incorrect answers is primarily driven by extreme attention events rather than typical behaviour. 

The bottom part of the figure plots the survival ECDFs for correct and incorrect generations and $p=99$. At each threshold $x$, the difference measures how much more or less likely correct generations are to exceed $x$ compared to incorrect generations. Negative values indicate that incorrect generations are more likely to exhibit extreme attention divergence. The darker areas around the line indicate the confidence intervals. 

Because the difference lies in the extremes and our method is probe dependent it is not possible to give a concrete example comparing the scores of a single truthful to a single false answer generation.

These results do not yet clarify where this uncertainty arises in the generated output. we analyse attention divergence at word level. Since transformers operate on subword units, we aggregate divergence values per word. Each word is then assigned to one of five semantic classes: named entities, numbers, stop words, punctuation, and any other words. 

Across the dataset, named entities exhibit higher average attention divergence than generic content words, while stop words and punctuation show lower divergence values. Although numeric tokens occur relatively infrequently, they display high divergence when they do appear. This is likely due to the nature of question answering datasets, which often require recalling specific years or quantities. 

This distinction becomes substantially clearer when focusing on extreme divergence events. We compute the 99th percentile of the full word distribution and assign all words whose divergence exceeds this threshold as belonging to the extreme tail. On the TriviaQA dataset, this analysis is based on 1445 generated words, of which 961 are classified as named entities, 435 as other content words, 33 as stop words, and 5 as numeric tokens. About 73\% of the words in the 99th percentile tail are named entities, while the remaining 27\% belongs to the "other" distribution. Stop words and punctuation are not in the tail. This indicates that our divergence signal is mostly localized at factual tokens. 

The most important observation is that there is cross dataset overlap and generalization. Given the total amount of heads in each model, this represents a small fraction ($\approx 0.5\%$) of the total. Indicating that the signal is not concentrated in a single universal "hallucination-" or "uncertainty-" aware attention head.
This result is consistent with earlier findings in literature suggesting that attention heads tend to specialize \citep{Zheng2024AttentionHO}. Other research by \citet{elhage2021mathematical} shows that even when individual heads have specialized roles, meaningful behaviour emerges from how multiple heads interact through the residual stream, instead of a single head. We observed a similar effect, while a few heads reoccur across datasets, most of our proposed signal comes from a larger set of spread out heads.

Although there is overlap across datasets, this does not imply sufficiency. No head appears across all four datasets, and many heads that are highly predictive in the lasso regularized probe are entirely absent in others. 

We have also compared the attention heads selected by the probe across all three model families. We consider an attention head to overlap across models if the exact layer $\times$ head pair is selected by the probe in at least one dataset for each model. We find that these overlaps are rare, and that only a small number of pairs occur across two different model families. No single attention head is shared across all three models. For example, a small number of heads overlap between Llama and Qwen, between Llama and Mistral, and between Qwen and Mistral. The complete data is in the appendix.

\begin{table}[t]
\centering
\small
\setlength{\tabcolsep}{6pt}
\renewcommand{\arraystretch}{1.1}
\begin{tabular}{lccc}
\toprule
Model & Early & Middle & Late \\
\midrule
Llama-3.2-3B & 15.5 & \textbf{53.5} & 31.0 \\
Qwen3-4B & 28.1 & \textbf{35.9} & 35.9 \\
Mistral-7B & 33.8 & \textbf{46.3} & 20.0 \\
\bottomrule
\end{tabular}
\caption{
Percentage of attention heads selected by the L1 probe (pooled across datasets)
located in early, middle, and late layers.
Layers are divided into equal thirds by depth.
}
\label{tab:layer_region_distribution}
\end{table}

\section{Discussion}
Our research investigates whether internal attention mechanisms in LLMs contain reliable signals of epistemic uncertainty, and whether these can be used for hallucination detection. Our results across multiple datasets, models, and ablation experiments provide evidence that attention divergence is predictive of answer correctness.

Our main finding is that attention divergence between attention maps and a uniform reference tends to be higher for incorrect or hallucinated outputs, particularly in the extreme tail of the distribution. This supports the hypothesis that hallucinations also arise from failures in internal computation.

Our ablation experiments provide insight into why this signal works. Removing individual attention heads that were identified by the probe has limited impact on performance. Removing entire groups of layers leads to a degradation in performance, specifically for the middle layers. This indicates there is a sparse subset of attention heads with an unclear relationship that are predictive of answer correctness.

Furthermore, the signal is concentrated on more semantically meaningful tokens, specifically, named entities and numerical values. Stop words and punctuation exhibit low divergence. This pattern matches recent literature showing that hallucination behaviour is related to factual content, rather than randomness. \citet{ferrando-2024} demonstrates that representations tied to entity knowledge are associated with whether a model correctly recalls facts. Likewise, \citet{ogasa-2025} highlights that failures in processing factual tokens such as numbers or entities can lead to breakdowns in reasoning, whereas this does not happen as much on generic tokens. 

A portion of the uncertainty signal is already present within the prompt. When attention divergence is computed solely over the prompt tokens, it remains predictive of answer correctness, even though the performance is weaker than over all answer tokens.

\section{Conclusion}

We proposed a simple, single-pass uncertainty measure based on the  (KL) Divergence between attention matrices and a uniform reference distribution representing maximum uncertainty. Across multiple datasets, task types, and model families, we showed that attention divergence is highly predictive of answer correctness and outperforms many existing baselines, while requiring negligible computational overhead.

Beyond performance, our analyses provide insight into where this uncertainty is located. We find that the proposed signal is concentrated in middle layers and at factual anchors such as named entities and numerical values.

Overall, our work suggests that internal attention dynamics contain reliable uncertainty signals that can be extracted with minimal computational cost. We believe that leveraging such white-box signals is a promising direction for improving the reliability and transparency of large language models.

\section{Limitations}

Despite the good results, there are a few important limitations to note.

First, while attention divergence is highly correlated with hallucinated and incorrect outputs, our work does not establish a causal relationship between specific patterns and false generations. The uncertainty signal is distributed across multiple heads and layers, and cannot be reduced to a simple logical rule such as "high divergence in a particular head implies an incorrect answer". As a result, our findings should be interpreted as a predictive signal. 

Secondly, although the lasso regularized probe identifies informative attention heads, the resulting model remains difficult to interpret. Consequently, the probe should not be viewed as an explanation of how uncertainty is represented, but rather as a tool for extracting the signal. Future research could try to set up a method that does not use any probe at all, by finding a relation as to which heads are predictive of hallucinations or answer correctness. 

Our model requires access to internal model architecture, with attention weights in particular, and therefore cannot be applied to black-box models that only give the generated text or output probabilities. However, this highlights the value of open and transparent model architectures. 

Although attention divergence itself is an informative signal, the specific combination of heads and layers selected by the probe varies across datasets and families. This suggests that a logistic regression probe might be too specific for uncertainty quantification.


\bibliography{custom}

\appendix

\section{Appendix}
\label{sec:appendix}

\subsection{Experimental Setup}

\subsubsection{Attention Extraction}

During greedy decoding, we extract attention weights on the full sequence (prompt plus generated tokens) with \texttt{output\_attentions=True} and \texttt{use\_cache=True}. We use the model’s returned post-softmax self-attention matrices for each generated token.

For each generated token $t$, and for each attention head $h$ in layer $l$ we obtain the attention distribution over all positions $\{1,...,t-1\}$. We measure how concentrated this is by computing its Kullback–Leibler (KL) divergence to a uniform distribution over the same $t+1$ positions. KL divergence is computed using natural logarithms, with a small $\epsilon$ value to clamp results for stability. 

For each head, the divergence values are averages over all generated token positions, resulting in a single scalar per head. Finally, all head features are concatenated into a feature vector $x \in \mathbb{R}^{L\times H}$, where $L$ is the number of layers and $H$ the number of heads per layer.

\subsubsection{TruthfulQA Answers}
For each example we use the MC1 (Multiple-choice 1) choice set from \texttt{mc1\_targets}. We randomly permute the choices per example, and define the correct answer as the index where the permuted \texttt{labels} equals 1. Since in the base dataset the correct answer is always the first one. The model is prompted to output a single letter (A, B, C, ...) (see Table ~\ref{tab:experimental-setup}). We extract the predicted letter and map it back to an index. Predictions without a letter are marked as incorrect.

\subsection{Directions for Future Research}
There are several interesting directions for future research building on our measure. First, while our results show that uncertainty lives strongest in the middle layers and is distributed across multiple heads, the underlying mechanisms remain unclear. Future work could focus on identifying groups of heads and layers than together encode uncertainty. Attention heads could be clustered based on similarity in their behaviour. Secondly, the current probing approach relies on linear logistic regression with lasso regularization. While this works great for selecting a spare subset of heads, future work could explore alternative probing methods that capture richer structure. 

Additionally, training probes on one dataset and evaluating them on others would allow for a better assessment on how uncertainty generalizes. Our token analysis indicates that attention divergence is mainly concentrated on semantically meaningful tokens, such as named entities, numbers, and dates. This suggests the possibility of localizing hallucinations within a generation rather than just scoring entire answers. Future work could try to compute attention divergence autoregressively to identify when a hallucination happens. This could allow models to highlight or flag specific parts of an output that are likely to be unreliable. Additionally, by comparing generations produced with and without retrieval augmented context (RAG), it would be possible to test whether divergence decreases when reliable external evidence is provided. Another direction for future research is evaluating attention divergence on a broader range of datasets and tasks types. In this paper, we mainly focus on question answering and reasoning, where correctness is easy to define. For instance, we did not include experiments for machine translation tasks due to the difficulty of defining and annotating hallucinations or factual errors in generated translations. 

Finally, attention divergence could potentially be used not only as a diagnostic signal (detecting hallucinations), but also as a training objective. A promising direction is to use reinforcement learning or other fine tuning approaches that penalize extreme attention divergence at critical tokens, such as named entities or numerical values. This may encourage the model to reduce hallucinations while still being fluent

\subsection{Data}

\subsubsection{Full Main Results}
Table \ref{tab:main_results} shows the full results of our experiment, including Expected Calibration Error (ECE) and accuracy along AUROC. 

\subsubsection{Sanity Check}

As said in section \ref{sec:sanity_checks} we performed several sanity checks to see whether our measure could be influenced by other factors, such as generation length. The results can be found below in table \ref{tab:sanity_checks}.

\begin{table}[h]
\centering
\small
\setlength{\tabcolsep}{6pt}
\renewcommand{\arraystretch}{1.1}

\begin{tabular}{lc}
\toprule
Baseline & AUROC \\
\midrule
Generation length        & 0.36 \\
Prompt length            & 0.44 \\
Raw output length        & 0.37 \\
Ends with punctuation    & 0.54 \\
Number of digits         & 0.48 \\
Generation length (perm.) & 0.50 \\
\bottomrule
\end{tabular}

\caption{
AUROC of sanity check features on TriviaQA using Llama-3.2-3B-Instruct.
All baselines are evaluated independently using a single random seed.
Permutation results are included as a sanity check.
}
\label{tab:sanity_checks}
\end{table}

\begin{table*}[t]
\centering
\small
\setlength{\tabcolsep}{6pt}
\renewcommand{\arraystretch}{1.1}

\begin{tabular*}{\textwidth}{@{\extracolsep{\fill}}llccc}
\toprule
Dataset & Model & AUROC & Accuracy & ECE \\
\midrule

\multirow{3}{*}{TruthfulQA}
& Llama-3.2-3B-Instruct   & $0.906 \pm 0.024$ & $\mathbf{0.838 \pm 0.029}$ & $\mathbf{0.201 \pm 0.012}$ \\
& Qwen3-4B-Instruct       & $\mathbf{0.906 \pm 0.025}$ & $0.827 \pm 0.026$ & $0.224 \pm 0.010$ \\
& Mistral-7B-Instruct-v0.2     & $0.891 \pm 0.024$ & $0.801 \pm 0.029$ & $0.219 \pm 0.017$ \\
\addlinespace

\multirow{3}{*}{TriviaQA}
& Llama-3.2-3B-Instruct   & $0.835 \pm 0.004$ & $\mathbf{0.791 \pm 0.002}$ & $\mathbf{0.059 \pm 0.005}$ \\
& Qwen3-4B-Instruct       & $\mathbf{0.846 \pm 0.001}$ & $0.770 \pm 0.005$ & $0.064 \pm 0.001$ \\
& Mistral-7B-Instruct-v0.2     & $0.835 \pm 0.004$ & $0.791 \pm 0.002$ & $\mathbf{0.059 \pm 0.005}$ \\
\addlinespace

\multirow{3}{*}{HotpotQA}
& Llama-3.2-3B-Instruct   & $\mathbf{0.8035 \pm 0.0219}$ & $0.7648 \pm 0.0140$ & $\mathbf{0.0644 \pm 0.0155}$ \\
& Qwen3-4B-Instruct       & $0.7963 \pm 0.0217$ & $\mathbf{0.7957 \pm 0.0225}$ & $0.0862 \pm 0.0050$ \\
& Mistral-7B-Instruct-v0.2     & $0.7782 \pm 0.0061$ & $0.7613 \pm 0.0078$ & $0.0957 \pm 0.0071$ \\
\addlinespace

\multirow{3}{*}{GSM8K}
& Llama-3.2-3B-Instruct   & $0.7672 \pm 0.0305$ & $0.7692 \pm 0.0137$ & $0.0601 \pm 0.0177$ \\
& Qwen3-4B-Instruct       & $\mathbf{0.9450 \pm 0.0151}$ & $\mathbf{0.9166 \pm 0.0132}$ & $0.0501 \pm 0.0078$ \\
& Mistral-7B-Instruct-v0.2     & $0.7889 \pm 0.0270$ & $0.7301 \pm 0.0129$ & $\mathbf{0.0439 \pm 0.0146}$ \\

\bottomrule
\end{tabular*}

\caption{
Results on validation set after training a lasso regularization probe to predict correctness from attention divergence mean pooled across the whole generation.
Within each dataset, the highest AUROC and accuracy and the lowest ECE are shown in bold.
Results are reported as mean $\pm$ standard deviation over three random seeds and 5 stratified cross-validation folds.
}
\label{tab:main_results}
\end{table*}

\onecolumn 

\begin{table}[h]
\centering
\begin{tabular}{p{0.35\linewidth} p{0.6\linewidth}}
\toprule
\textbf{Parameter} & \textbf{Value} \\
\midrule
\multicolumn{2}{l}{\textit{Models and evaluation}} \\
Models & Llama-3.2-3B-Instruct, Mistral-7B-Instruct-v0.2, Qwen3-4B-Instruct-2507 \\
ECE bins & 10 \\
\midrule
\multicolumn{2}{l}{\textit{Probe and training}} \\
Probe & Logistic regression (L1 / lasso) \\
Solver & \texttt{saga} \\
Regularization strength ($C$) & 1.0 \\
Max probe iterations & 20{,}000 \\
Train / validation split & 80\% / 20\% (stratified) \\
Cross-validation & 5-fold stratified \\
Random seeds & 3 \\
\midrule
\multicolumn{2}{l}{\textit{Generation and attention features}} \\
Decoding & Greedy ($T=0$, no sampling) \\
Max new tokens & 32 (HotpotQA, TriviaQA), 8 (TruthfulQA), 256 (GSM8K)\\
Attention aggregation & Mean over generated tokens \\
\midrule
\multicolumn{2}{l}{\textit{Datasets}} \\
GSM8K & 1319 samples; numeric answer extraction \\
TriviaQA & 2000 samples; substring match against aliases \\
HotpotQA & 2000 samples; context truncated to 2048 tokens \\
TruthfulQA & 272 samples per seed; MC1 with shuffled choices \\
\midrule
\multicolumn{2}{l}{\textit{Prompt per Dataset}} \\
GSM8K & \textit{Solve the problem and give the final numeric answer.} \\
TriviaQA & \textit{Answer the question briefly.} \\
HotpotQA & \textit{Answer using the context. Be brief.} \\
TruthfulQA & \textit{Answer with ONLY the letter (A, B, C, ...).} \\

\bottomrule
\caption{Experimental setup and hyperparameters. All settings apply to Llama, Mistral, and Qwen unless otherwise stated.}
\label{tab:experimental-setup}
\end{tabular}

\end{table}

\setlength{\LTleft}{\fill}
\setlength{\LTright}{\fill}

\begin{longtable}{lccccccl}
\label{tab:l1_all_models_long} \\
\toprule
Model & Layer & Head & GSM8K & TruthfulQA & TriviaQA & HotpotQA & Total \\
\midrule
\endfirsthead

\toprule
Model & Layer & Head & GSM8K & TruthfulQA & TriviaQA & HotpotQA & Total \\
\midrule
\endhead

\midrule
\multicolumn{8}{r}{\emph{Continued on next page}}
\endfoot

\endlastfoot

\textbf{Llama} & 21 & 02 & 10 & -- & -- & 6 & 16 \\
& 13 & 18 & -- & 10 & 9 & -- & 19 \\
& 17 & 06 & 9 & -- & -- & -- & 9 \\
& 21 & 13 & 9 & -- & -- & -- & 9 \\
& 10 & 10 & -- & 10 & -- & -- & 10 \\
& 12 & 06 & -- & 9 & -- & -- & 9 \\
& 13 & 21 & -- & 9 & -- & -- & 9 \\
& 14 & 06 & -- & 9 & -- & -- & 9 \\
& 14 & 10 & -- & 9 & -- & -- & 9 \\
& 15 & 22 & -- & 9 & -- & -- & 9 \\
& 23 & 15 & -- & 9 & -- & -- & 9 \\
& 25 & 01 & -- & 9 & -- & -- & 9 \\
& 03 & 09 & 8 & -- & -- & -- & 8 \\
& 04 & 05 & 8 & -- & -- & -- & 8 \\
& 20 & 08 & 8 & -- & -- & -- & 8 \\
& 07 & 18 & 8 & -- & -- & -- & 8 \\
& 21 & 05 & -- & -- & -- & 8 & 8 \\
& 08 & 13 & -- & -- & -- & 7 & 7 \\
& 08 & 22 & -- & -- & -- & 7 & 7 \\
& 09 & 19 & -- & -- & -- & 7 & 7 \\
& 18 & 09 & -- & -- & -- & 7 & 7 \\
& 06 & 21 & 7 & -- & -- & -- & 7 \\
& 16 & 09 & 7 & -- & -- & -- & 7 \\
& 19 & 20 & 7 & -- & -- & -- & 7 \\
& 24 & 12 & 7 & -- & -- & -- & 7 \\
& 04 & 17 & 7 & -- & -- & -- & 7 \\
& 18 & 12 & -- & -- & -- & 6 & 6 \\
& 08 & 00 & -- & -- & -- & 6 & 6 \\
& 14 & 13 & -- & -- & -- & 6 & 6 \\
& 27 & 05 & -- & -- & -- & 6 & 6 \\
& 08 & 14 & -- & -- & 6 & 6 & 12 \\
& 12 & 01 & -- & -- & -- & 6 & 6 \\

\midrule
\textbf{Qwen} & 22 & 02 & -- & 10 & -- & -- & 10 \\
& 26 & 05 & -- & 10 & -- & -- & 10 \\
& 22 & 00 & -- & 9 & -- & -- & 9 \\
& 28 & 15 & -- & 8 & -- & -- & 8 \\
& 31 & 08 & -- & 8 & -- & -- & 8 \\
& 20 & 04 & -- & 6 & -- & 8 & 14 \\
& 25 & 05 & -- & -- & -- & 8 & 8 \\
& 22 & 08 & -- & -- & -- & 7 & 7 \\
& 23 & 12 & -- & -- & -- & 7 & 7 \\
& 25 & 08 & -- & -- & -- & 7 & 7 \\
& 30 & 10 & -- & -- & 6 & -- & 6 \\
& 14 & 05 & -- & -- & 6 & -- & 6 \\
& 00 & 13 & -- & 6 & -- & -- & 6 \\
& 23 & 05 & -- & 6 & -- & -- & 6 \\
& 27 & 05 & -- & 6 & -- & -- & 6 \\
& 28 & 07 & -- & 6 & -- & -- & 6 \\
& 30 & 00 & -- & 6 & -- & -- & 6 \\
& 29 & 11 & -- & 6 & -- & -- & 6 \\
& 00 & 01 & 4 & -- & -- & 5 & 9 \\
& 00 & 12 & 4 & -- & -- & -- & 4 \\

\midrule
\textbf{Mistral} & 31 & 02 & 10 & -- & 6 & 8 & 24 \\
& 14 & 08 & -- & 10 & -- & 7 & 17 \\
& 14 & 10 & -- & 10 & -- & -- & 10 \\
& 31 & 03 & -- & 6 & 10 & -- & 16 \\
& 10 & 23 & -- & 9 & -- & -- & 9 \\
& 15 & 18 & -- & 9 & -- & -- & 9 \\
& 16 & 20 & -- & 9 & -- & -- & 9 \\
& 16 & 29 & -- & 9 & -- & -- & 9 \\
& 30 & 07 & 7 & -- & -- & -- & 7 \\
& 01 & 30 & 7 & -- & -- & -- & 7 \\
& 06 & 19 & 7 & -- & -- & -- & 7 \\
& 11 & 22 & -- & -- & -- & 7 & 7 \\
& 21 & 04 & -- & -- & -- & 7 & 7 \\
& 18 & 24 & -- & -- & -- & 7 & 7 \\
& 05 & 26 & -- & -- & -- & 6 & 6 \\
& 07 & 18 & -- & -- & -- & 6 & 6 \\
& 12 & 17 & -- & -- & -- & 6 & 6 \\
& 16 & 01 & -- & -- & -- & 6 & 6 \\
& 18 & 23 & -- & -- & 6 & 6 & 12 \\

\caption{All attention heads selected by the L1 probe across models, datasets, and random seeds. Values denote the number of seeds (out of 10) in which a head was selected. A dash (--) indicates that the head was not selected for that dataset.}

\end{longtable}

\twocolumn

\end{document}